\documentclass[runningheads]{llncs}
\usepackage{graphicx}
\usepackage{amssymb}
\usepackage{amsmath}
\usepackage[pagebackref,breaklinks,colorlinks]{hyperref}
\usepackage[capitalize]{cleveref}
\usepackage{booktabs}
\usepackage[table]{xcolor}
\usepackage{subfig}
\usepackage{bm}

\crefname{section}{Sec.}{Secs.}
\Crefname{section}{Section}{Sections}
\Crefname{table}{Table}{Tables}
\crefname{table}{Tab.}{Tabs.}

\usepackage{multirow}
\usepackage{booktabs} 
\usepackage{tabularx} 
\usepackage{makecell} 
\usepackage{arydshln} 
\usepackage{amsmath}
\usepackage[symbol]{footmisc}

\usepackage{booktabs} 
\begin{document}
\title{PathInsight: Instruction Tuning of Multimodal Datasets and Models for Intelligence Assisted Diagnosis in Histopathology}
\titlerunning{PathInsight: Instruction Tuning of Multimodal Datasets and Models}

\author{Xiaomin Wu$^{\ast 1}$ \and
 Rui Xu$^{\ast 1}$ \and
 Pengchen Wei$^{\ast 1}$ \and
Wenkang Qin$^{1}$ \and Peixiang Huang$^{1}$\and Ziheng Li$^{1}$\and Lin Luo$^{\dag 1}$}

\institute{ $^{1}$ College of Engineering,  Peking University, Beijing, China}

\footnotetext[1]{These authors contributed equally to this work.} \footnotetext[2]{Corresponding author.}


\authorrunning{X.Wu et al.}


\maketitle
\begin{abstract}
Pathological diagnosis remains the definitive standard for identifying tumors. The rise of multimodal large models has simplified the process of integrating image analysis with textual descriptions. Despite this advancement, the substantial costs associated with training and deploying these complex multimodal models, together with a scarcity of high-quality training datasets, create a significant divide between cutting-edge technology and its application in the clinical setting. We had meticulously compiled a dataset of approximately 45,000 cases, covering over 6 different tasks, including the classification of organ tissues, generating pathology report descriptions, and addressing pathology-related questions and answers. We have fine-tuned multimodal large models, specifically LLaVA, Qwen-VL, InternLM, with this dataset to enhance instruction-based performance. We conducted a qualitative assessment of the capabilities of the base model and the fine-tuned model in performing image captioning and classification tasks on the specific dataset. The evaluation results demonstrate that the fine-tuned model exhibits proficiency in addressing typical pathological questions. We hope that by making both our models and datasets publicly available, they can be valuable to the medical and research communities.

\end{abstract}
\section{Introduction}
Pathological examination remains the cornerstone of tumor and cancer diagnostics, serving as the gold standard for diagnostic techniques~\cite{rosai2007microscopy}. The meticulous analysis of tissue samples under the microscope by experienced pathologists enables the detection, classification, and grading of cancerous lesions, providing critical insights into the most appropriate therapeutic strategies. However, there is a relative scarcity of senior pathologists, and the processes of accessing and consulting pathological knowledge are often cumbersome. Additionally, the procedures of slide examination, diagnosis, and meticulous documentation are time-consuming, further exacerbating the complexity and workload inherent in the diagnostic process~\cite{asare2002improving}.

The emergence of digital tools such as Whole Slide Imaging (WSI) has significantly facilitated the process of pathological diagnosis, making the storage and transfer of pathological data easier than ever~\cite{farahani2015whole,hanna2022integrating}. Certainly, a substantial array of sophisticated methodologies pertaining to Whole Slide Imaging (WSI) within the realm of digital pathology has come to the forefront in recent scholarly discourse~\cite{xu2023scaat,huang2023assessing,Qin_2022_ACCV}. Concurrently, large language models are increasingly being harnessed for their capabilities in machine learning and deep learning algorithms, particularly in the analysis of pathological images~\cite{rahaman2020survey}. Some state-of-the-art methods have become extensively utilized for tasks including disease classification~\cite{yan2020breast,echle2021deep}, segmentation of pathological lesions~\cite{hemelings2021pathological}, and prognostication of patient~\cite{sandroni2021brain}. However, their efficacy is often contingent on the availability of large sets of annotated data for training. Additionally, these tasks are generally performed in isolation, and their application domains tend to be specialized.

Currently, there exists a significant gap in the availability of an integrated, comprehensive, and user-friendly algorithmic model that encapsulates the essential capabilities for pathologists or medical students engaged in diagnosis or educational activities. Such capabilities encompass image analysis, resolution of pathology-related image queries, retrieval of pathological knowledge, and the generation of complete diagnostic reports.

Recently, Large Language Models (LLMs), epitomized by GPT ~\cite{floridi2020gpt} have ignited a surge of research interest. In particular, the emergence of multimodal large language models, such as GPT-4Vision (GPT4V)~\cite{openai2023gpt}, LLaVA~\cite{liu2024visual}, Qwen-VL~\cite{bai2023qwen}, InternLM~\cite{zhang2023internlm}, and CogVLM~\cite{wang2023cogvlm}, has facilitated the seamless integration of image and text data. These models are capable of concurrently addressing a variety of tasks, such as text recognition, visual reasoning, visual question answering, and caption generation ~\cite{fu2023mme}. In clinical practice, these models are envisioned to synergize with pathologists' queries to analyze medical images, thereby offering broader knowledge, more inspiration, deeper diagnostic insights, and accelerated diagnostic processes.
In the context of pathology education, these models are envisioned to assist students in better understanding the details of pathological images and the corresponding medical knowledge.

Smart Doctor models typically restrict interactions to text communication, with data often sourced from quickly established question-and-answer sessions between patients and doctors with simple conditions. 
However, pathological diagnosis is extremely specialized, which presents unique challenges that these text-based models~\cite{wang2023huatuo} are not equipped to address.
Current multimodal large models primarily focus on general domains, emphasizing tasks related to image perception and cognition but lacking in domain-specific knowledge and capabilities for the medical field. 
The models like LLaVA-Med~\cite{li2024llava} and PathAsst ~\cite{Sun2023Pathasst} represent initial efforts to integrate multimodal large language models with medical applications.
However, LLaVA-Med~\cite{li2024llava} mainly concentrates on medical imaging modalities such as CT and MR, with a lesser emphasis on pathological images, which may result in relatively weaker pathological capabilities compared to other data types. PathAsst~\cite{Sun2023Pathasst}, on the other hand, serves as a foundational model tailored for pathology, encompassing a variety of pathological tasks. However, its development necessitates significant investment in data, resources, and human expertise.

To address these challenges, we propose a pathology-focused multimodal instruction fine-tuning dataset. By fine-tuning existing widely acknowledged multimodal models with this dataset, we aim to achieve practical and user-friendly multimodal models for pathology. Our main contributions are as follows:
\begin{itemize}

\item We constructed a pathology-oriented dataset that encompasses a range of tasks including pathological image grading, classification typing, pathological image caption generation, image descriptions, image-based question-answering, and dialogue interactions, comprising approximately 45,000 instances.

\item Utilizing this pathology dataset, we have fine-tuned a selection of widely recognized multimodal models, employing strategies including Low-Rank Adaptation (LoRA) and full-parameter tuning. This enhances the practical application of multimodal models in the field of pathology by augmenting multi-tasking capabilities and supporting extensive and comprehensive uses.

\item Quantitative evaluations of model's capabilities have been undertaken in tasks, such as classification, image caption generation, and question-answering. Additionally, we invited experienced pathologist to provide qualitative assessments of the model's honesty in answering open-ended questions. The results indicate that our constructed dataset and fine-tuning paradigm enable effective handling of tasks like classification and caption generation, and also demonstrate potential in dealing with open-ended, interpretive questions that pathologist encounter in clinical practice.
\end{itemize}

\begin{figure}[!t]
\includegraphics[width=\textwidth]{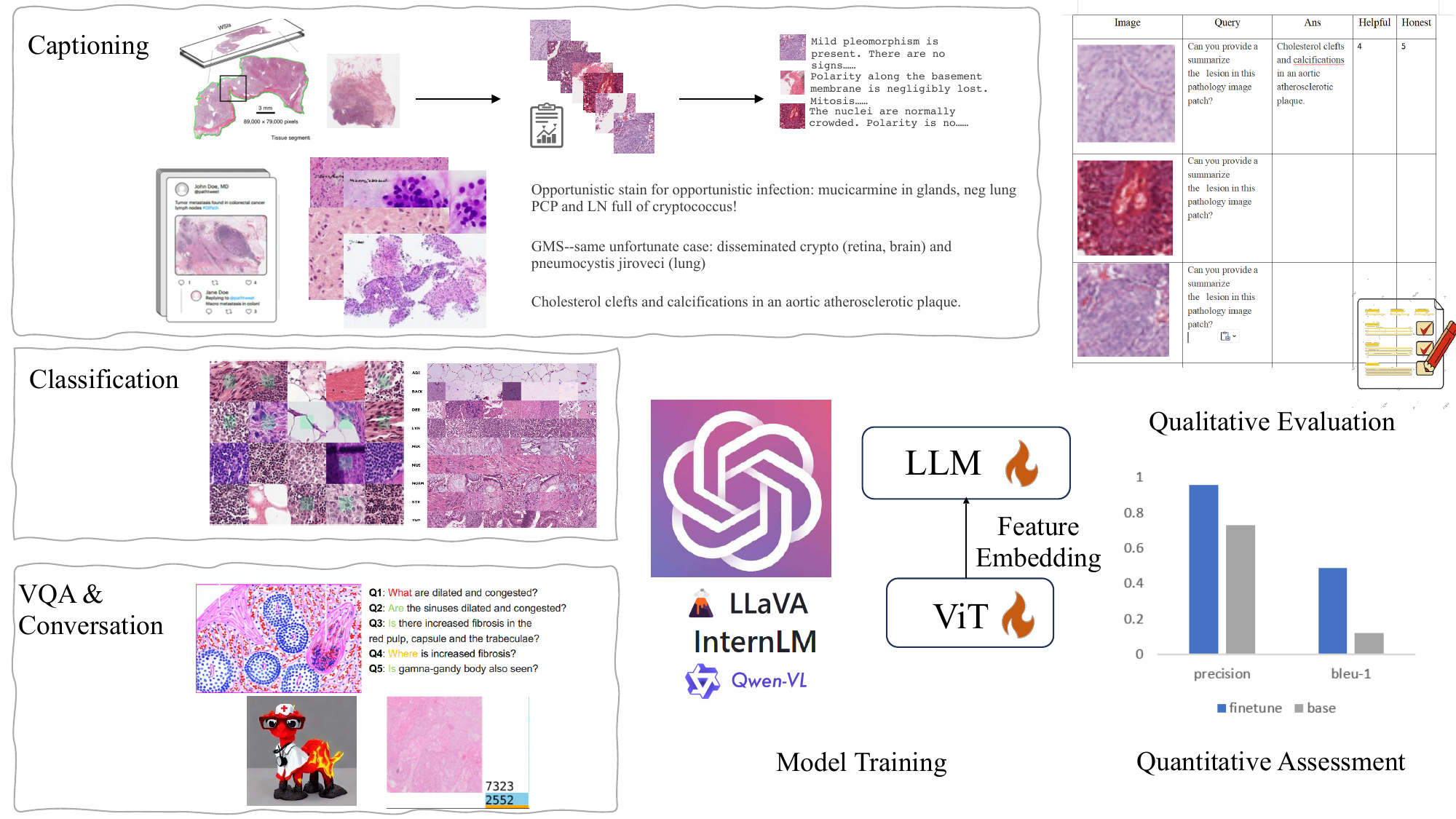}
\caption{Instruction tuning of multimodal dataset and models for intelligent assisted
diagnosis in histopathology.} \label{fig1}
\end{figure}

\section{Methods}
\subsection{PathEnhanceDS Compiling}

To ensure the comprehensiveness and diversity of our dataset, meticulous considerations were undertaken regarding the assortment and synthesis of data sources, which culminated in the creation of a dataset named $\textbf{PathEnhanceDS}$, with the specifics delineated as follows:

\begin{figure}[t!]
\includegraphics[width=\textwidth]{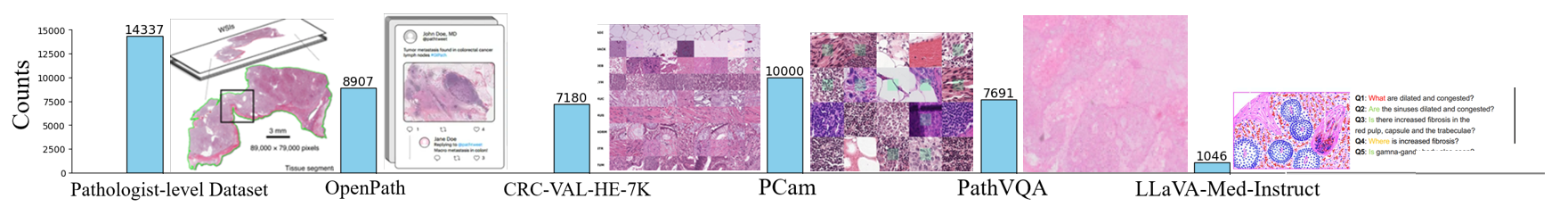}
\caption{The composition of PathEnhanceDS.}\label{fig2}
\end{figure}

Initially, we meticulously curated datasets from both cutting-edge and established data repositories to encompass a wide array of pathologies and diagnostic scenarios. The selection process was governed by several pivotal criteria: the relevance to diagnostic pathology, the quality of annotations, and the compatibility across different tasks such as Captioning, Classification, Visual Question Answering (VQA), and Conversation.

\begin{table}[t!]
\renewcommand{\arraystretch}{1.2}
\centering
\caption{Data retention percentages after cleaning for various datasets}
\label{table:retention per}
\begin{tabularx}{\textwidth}{@{}Xlcccc@{}}
\toprule
\textbf{Task} & \textbf{Count} & \textbf{Dataset} & \textbf{Original} & \textbf{Cleaned} & \textbf{Remaining} \\ 
\midrule
\multirow{2}{*}{\makecell[lt]{Captioning}}       & \multirow{2}{*}{23.2K} & Pathologist-level Dataset & 21.2K  & 14.3K & 66.7\% \\
                                                  &                      & OpenPath                  & 208K   & 8.9K  & 4.3\% \\
\hdashline
\multirow{2}{*}{\makecell[lt]{Classification}} & \multirow{2}{*}{17.2K} & PCam                     & 328K   & 10K   & 3.0\% \\
                                                &                      & CRC-VAL-HE-7K             & 7.2K   & 7.2K  & 100\% \\
\hdashline
\multirow{1}{*}{\makecell[lt]{VQA}}           & \multirow{1}{*}{1.0K} & PathVQA                   & 5.0K   & 1.0K  & 20\% \\
\hdashline
\multirow{2}{*}{\makecell[lt]{Conversation}}  & \multirow{2}{*}{7.6K} & LLaVA-Med-Instruct-10k  & 10K    & 2.5K  & 25\% \\
                                              &                      & LLaVA-Med-Instruct-60k  & 60K    & 5.2K  & 8.7\% \\
\bottomrule
\end{tabularx}
\end{table}


We relied on the Pathologist-level Dataset~\cite{zhang2019pathologist}, a collection of data annotated by medical professionals that provided patch-level descriptive information. We retained the initial diagnostic descriptions provided by seasoned pathologists to ensure the textual data mirrored the caliber of professional expertise.

Further, to expand our dataset's heterogeneity, we incorporated OpenPath~\cite{huang2023visual}, derived from the broad-reaching social media platform Twitter, which offered an unconventional perspective on pathology data through publicly accessible Twitter IDs. The preeminent 20,000 samples exhibiting high semantic congruity between images and their captions based on the BiomedCLIP~\cite{zhang2024biomedclip} were selected, with an emphasis on HE stained images.

To ensure balanced binary classification, we integrated the eminent PCam~\cite{Veeling2018-qh}, recognized for its binary classification challenges within the pathology domain. Additionally, the CRC-VAL-HE-7K~\cite{Kather2018Histological}, focused on multi-classification of tissue organs, was selected for its rigorous curation and substantial relevance to the field.

The adoption of PathVQA~\cite{he2020pathvqa} was motivated by its comprehensive compilation of pathology images paired with question-answer pairs, sourced from a wide gamut of diagnostic categories. This vital component ensured that our dataset was not only rich in visual content but also in the interactive, interrogative dialogue that is essential for training sophisticated Visual Question Answering models. Inclusivity of PathVQA allowed us to simulate a more interactive environment, closely imitating the dynamic interplay between a pathologist and their diagnostic workflow. We incorporated images from this dataset using the predictive capabilities of the BiomedCLIP~\cite{zhang2024biomedclip} to ensure a focus on HE histopathology, enriching the dataset with structured pathological Q\&A content.

LLaVA-Med-Instruct~\cite{li2024llava-med}, on the other hand, was selected for its novel conversational style format data. This dataset, characterized by its conversation-based queries that necessitate thorough internalization and replication of diagnostic thought processes, provided an exceptional opportunity to imbue our dataset with the finesse of clinical judgment and decision-making. Dialogue data pertaining exclusively to pathology and consisting of 7,691 multi-turn dialogues with 28,710 question-answer pairs were included to capture the interactive aspect of pathology diagnostics.


In conjunction with the corresponding queries, we transform the relevant data into instructional data and format it as ChatML. Given a pathological image $P_v$ and its corresponding description $P_d$, we generate a query $P_q$ that solicits a description of the pathological image. For the captioning task, the query here is randomly extracted from a series of questions. Specific prompts (queries) can be found in the Appendix.The arrangement of ($P_v$, $P_d$, $P_q$) forms a single-round interaction in the following format:

\begin{itemize}
\item User: \texttt{<img>$P_v$</img> $P_q$<STOP>}
\item Assistant: \texttt{$P_d$<STOP>}
\end{itemize}

Our diverse choices of datasets were instrumental in constructing an effective and wide-reaching study, shedding light on the multifaced realm of pathological multimodal language models.

\subsection{PathoSync Tuning}
After constructing PathEnhanceDS instruction tuning dataset, the primary challenge we faced was adapting general-purpose models to effectively address specific issues within the medical histopathology domain. 

We experimented with three representative multimodal large language models: LLaVA~\cite{liu2024visual}, Qwen-VL~\cite{bai2023qwen}, InternLM~\cite{zhang2023internlm}, and  which are recognized for their exceptional performance in natural visual language domains.

LLaVA: Based on the LLaMA language model, LLaVA employs a preceding vision encoder (based on the CLIP model) to convert images into patch features.

Qwen-VL-7B: Qwen-VL-7B integrated the ViT-bigG~\cite{ilharco_gabriel_2021_5143773} project as its vision encoder and introduced a single randomly initialized cross-attention layer to facilitate close interaction between textual and visual information.

InternLM: InternLM-XComposer2 is a groundbreaking vision-language large model (VLLM) based on InternLM2-7B excelling in free-form text-image composition and comprehension

The pre-training phase of these models mainly focused on tasks such as image comprehension, text recognition, and visual reasoning, with minimal exposure to medical or pathological data. Consequently, there is a significant disparity between our dataset and the data the models were originally exposed to. Notably, medical images often contain complex pathological features, posing a higher demand on the models' visual understanding capabilities.

To bridge this gap and enhance the models' understanding of medical visual information, we fully enabled the fine-tuning of the Vision Transformer (ViT) component during training. During the fine-tuning process, we fine-tuned both the vision encoder (ViT/CLIP) and the language model parameters, with full parameters or LoRA parameters update. 
Fine-tuning the entire set of parameters allows the model to learn information between pathological images and text to the greatest extent possible, but often requires longer training times and resource consumption. We also experiments using the Low-Rank Adaptation (LoRA)~\cite{hu2021lora} training method. This involves freezing the parameters of the pre-trained model and introducing a trainable low-rank separable matrix, referred to as the side-path matrix, in each layer of the Transformer. By adding the side-path output to the initial path output, a new path is formed as input to the network. During training, only these newly introduced side-path matrix parameters are updated, thereby speeding up the training process and reducing the forgetting of prior knowledge when learning pathological information.

While adhering to the fundamental training strategy, we ensured that the sequence length was set to 2048. For the fine-tuning dataset, text was firstly encoded using a tokenizer, and images paths were extracted accordingly. In the vision encoder, the images underwent feature extraction, after which the features were downsampled to a 256-dimensional representation to be combined with text tokens. We set the learning rate as 1e-5, batch size as 2, and the warm-up ratio to 0.05 to increase the learning rate smoothly, reducing drastic fluctuations in model parameters during the initial phase of training and facilitating stable convergence. We trained the model on 45,000+ training data points for 3 epochs, and saved checkpoints for subsequent model evaluation.

\section{Experiments}
After fine-tuning the model with PathEnhanceDS, it was imperative to rigorously assess its performance within the domain of pathological medicine. The primary aim of our experimental design is to investigate the feasibility of transferring large language multimodal models, which have been pretrained in the natural domain, to the realm of pathology by instructional fine-tuning, and to verify the effectiveness of the multimodal pathology dataset we have constructed.

\subsection{Evaluation Metrics} 
In order to verify the effectiveness of fine-tuning, the evaluation focused on several core tasks defined earlier, classification of pathological images (identifying the presence of pathological lesions and the type of tissue involved), the capability to generate captions for pathological images, and visual reasoning abilities based on pathological imagery.

 For classification tasks, particularly for identification of tissue types, precision, recall, and F1 score have been chosen as the main evaluation metrics. 

In open-ended generative tasks, such as image captioning, due to the unpredictable nature of the task, traditional machine translation evaluation metrics, such as BLEU and ROUGE, are used to assess the consistency between generated texts and reference texts. BLEU-1 is used to measure the level of exact word-level matches between the generated text and the reference text, namely the overlap of 1-grams, which reflects the accuracy of word selection. The ROUGE metric, on the other hand, measures the frequency with which reference text words appear in the generated text, evaluating the comprehensiveness of the generated text from the perspective of recall. Together, these two metrics provide a quantitative means to assess the performance of models in generative tasks.

\subsection{Evaluation Results} 
The results indicate that after fine-tuning, the model has significantly improved its relevance. Notably, In CRC classification tasks, the performance is highly consistent with the ground truth (GT). The caption generated by the model are structurally accurate and generally conform to the fundamental knowledge of medical pathology in terms of content. However, there remain some discrepancies in the accuracy of details.

In the quantitative tasks, we further calculated the answer precision in CRC classification table \ref{tab:quantitative_results}. For PCAM classification tasks, Acc was improved from 0.5 to 0.96 by fine-tuning. The base model exhibited good precision on the CRC classification dataset, but the recall and f1-score ware low. The f1-score in CRC classification task of Qwen-VL model increased from 0.138 to about 0.978. After fine-tuning, the LLaVa model showed improved results, with the recall increasing from 0.008 to 0.967.

We also quantitatively assessed the performance of the captioning task \ref{tab:quantitative_results}. Despite targeted optimization for large language model generation tasks, significant differences from traditional machine translation tasks remain. Specifically, for the image captioning task on the Pathologist-level Dataset, the Qwen-VL
model's BLEU-1 score improved from 0.09 to 0.41, Rouge-1 score improved from 0.15 to 0.58. The LLaVA-1.5
model's BLEU-1 score improved from 0.099 to 0.421 (Full parameters fine-tuning), 0.371 (LoRA parameters fine-tuning), Rouge-1 score improved from 0.145 to 0.583 (Full parameters fine-tuning), 0.539 (LoRA parameters fine-tuning), indicating a substantial enhancement in performance. Additionally, for the OpenPath dataset captioning, we calculated BLEU-1 and ROUGE-1 scores. However, given the high degree of individual variability in data from Twitter, these conventional metrics may not be entirely suitable for assessing model performance on such heterogeneous data sources. After removing the OpenPath, we applied both LoRA and full-parameter training to the LLaVA model.
To visually demonstrate the model's performance, we presents some examples of the model's responses in Appendix.


\begin{table}[t!]
    \centering
    \setlength{\tabcolsep}{4pt} 
    \caption{The quantitative results before and after fine-tuning models. $\thicksim$ means without fine-tuning.}
    \label{tab:quantitative_results}
    \renewcommand{\arraystretch}{1.2}
  \begin{tabular}{@{}l|cc|cc|cc@{}}
    \bottomrule[1.2pt]
    \multirow{2}{*}{Model} & \multirow{2}{*}{OpenPath}&\multirow{2}{*}{LoRA}  & \multicolumn{2}{c|}{CRC }&\multicolumn{2}{c}{Pathologist-level} \\ 
    &&& F1-Score & Recall&BLEU-1 & Rouge-1 \\ 
    \midrule[1.2pt]
    LLaVA-1.5 & \multicolumn{2}{c|}{$\thicksim$} & 0.016 & 0.008 & 0.099 & 0.145 \\
    \hdashline
    \multirow{4}{*}{LLaVA-1.5-PathInsight} && & 0.991 & 0.983   & 0.382 & 0.555 \\
     &\checkmark& & 0.983 & 0.967 &0.421 & 0.583 \\
    & &\checkmark  & 0.987 & 0.975  & 0.401 & 0.566 \\
    &\checkmark&\checkmark  & 0.991 & 0.983 & 0.371 & 0.539 \\
    \hdashline
    Qwen-VL-Chat & \multicolumn{2}{c|}{$\thicksim$} & 0.138 & 0.074 & 0.095 & 0.155 \\
    \hdashline
    Qwen-VL-PathInsight &\checkmark&  & 0.978 & 0.958 & 0.410 & 0.579 \\
    \hdashline
    InternLM  & \multicolumn{2}{c|}{$\thicksim$} & 0.064 & 0.033 & 0.098 & 0.149 \\
    \hdashline
    InternLM-PathInsight &\checkmark&  & 0.933 & 0.876 & 0.398 & 0.558 \\
    \bottomrule[1.2pt]
  \end{tabular}
\end{table}

\section{Conclusion}
The integration of multimodal large models has significantly improved the handling of diverse tasks in a unified framework, enhancing both image and text analysis. Especially in computational pathology, these models enable clinicians and medical students to access and document diagnostic information more efficiently. However, the application in medical pathology is limited by high data acquisition costs and the scarcity of quality datasets.

To overcome these challenges, we developed PathEnhanceDS, a comprehensive dataset with over 45,000 cases that include disease grading, tissue classification, and pathology report generation tasks. By fine-tuning models like Qwen-VL, InternLM, and LLaVA, our evaluations show significant performance improvements in pathology, adhering to clinical medicine standards for open-ended questions. This highlights the effectiveness of our fine-tuning approach and the adaptability of our carefully curated dataset.


\bibliographystyle{splncs04}
\bibliography{ref}


\end{document}